\PassOptionsToPackage{unicode}{hyperref}
\PassOptionsToPackage{hyphens}{url}
\documentclass[
]{article}
\usepackage{amsmath,amssymb}
\usepackage{lmodern}
\usepackage{iftex}
\ifPDFTeX
  \usepackage[T1]{fontenc}
  \usepackage[utf8]{inputenc}
  \usepackage{textcomp} 
\else 
  \usepackage{unicode-math}
  \defaultfontfeatures{Scale=MatchLowercase}
  \defaultfontfeatures[\rmfamily]{Ligatures=TeX,Scale=1}
\fi
\IfFileExists{upquote.sty}{\usepackage{upquote}}{}
\IfFileExists{microtype.sty}{
  \usepackage[]{microtype}
  \UseMicrotypeSet[protrusion]{basicmath} 
}{}
\makeatletter
\@ifundefined{KOMAClassName}{
  \IfFileExists{parskip.sty}{%
    \usepackage{parskip}
  }{
    \setlength{\parindent}{0pt}
    \setlength{\parskip}{6pt plus 2pt minus 1pt}}
}{
  \KOMAoptions{parskip=half}}
\makeatother
\usepackage{xcolor}
\usepackage[margin=1in]{geometry}
\usepackage{color}
\usepackage{fancyvrb}

\DefineVerbatimEnvironment{Highlighting}{Verbatim}{commandchars=\\\{\}}
\usepackage{framed}
\definecolor{shadecolor}{RGB}{248,248,248}
\newenvironment{Shaded}{\begin{snugshade}}{\end{snugshade}}

\newcommand{\AttributeTok}[1]{\textcolor[rgb]{0.77,0.63,0.00}{#1}}

\newcommand{\BuiltInTok}[1]{#1}

\newcommand{\CommentTok}[1]{\textcolor[rgb]{0.56,0.35,0.01}{\textit{#1}}}

\newcommand{\DataTypeTok}[1]{\textcolor[rgb]{0.13,0.29,0.53}{#1}}

\newcommand{\ExtensionTok}[1]{#1}

\newcommand{\FunctionTok}[1]{\textcolor[rgb]{0.00,0.00,0.00}{#1}}

\newcommand{\KeywordTok}[1]{\textcolor[rgb]{0.13,0.29,0.53}{\textbf{#1}}}
\newcommand{\NormalTok}[1]{#1}
\newcommand{\OperatorTok}[1]{\textcolor[rgb]{0.81,0.36,0.00}{\textbf{#1}}}

\newcommand{\PreprocessorTok}[1]{\textcolor[rgb]{0.56,0.35,0.01}{\textit{#1}}}

\newcommand{\StringTok}[1]{\textcolor[rgb]{0.31,0.60,0.02}{#1}}
\newcommand{\VariableTok}[1]{\textcolor[rgb]{0.00,0.00,0.00}{#1}}

\usepackage{graphicx}
\makeatletter
\def\maxwidth{\ifdim\Gin@nat@width>\linewidth\linewidth\else\Gin@nat@width\fi}
\def\maxheight{\ifdim\Gin@nat@height>\textheight\textheight\else\Gin@nat@height\fi}
\makeatother
\setkeys{Gin}{width=\maxwidth,height=\maxheight,keepaspectratio}
\makeatletter
\def\fps@figure{htbp}
\makeatother
\providecommand{\tightlist}{%
  \setlength{\itemsep}{0pt}\setlength{\parskip}{0pt}}
\usepackage{amsmath}
\usepackage{eufrak}
\usepackage{float}
\usepackage{xcolor}
\usepackage[bookmarks,bookmarksnumbered, pdfborder={0 0 0},linktocpage, colorlinks=true]{hyperref}
\hypersetup{citecolor={blue}}
\ifLuaTeX
  \usepackage{selnolig}  
\fi

\usepackage{graphicx}

\def\BibTeX{{\rm B\kern-.05em{\sc i\kern-.025em b}\kern-.08em
    T\kern-.1667em\lower.7ex\hbox{E}\kern-.125emX}}
    
\IfFileExists{bookmark.sty}{\usepackage{bookmark}}{\usepackage{hyperref}}
\IfFileExists{xurl.sty}{\usepackage{xurl}}{} 
\hypersetup{
  pdftitle={Prototyping Vehicle Control Applications using CAT Vehicle Simulator},
  pdfauthor={Rahul Bhadani, Jonathan Sprinkle},
  pdfkeywords={Autonomous Vehicles, Control Systems, Software,
Model-based Design, ROS, Robotics, Sensors},
  hidelinks,
  pdfcreator={LaTeX via pandoc}}

\title{Prototyping Vehicle Control Applications\\Using the CAT Vehicle Simulator}
\author{Rahul Bhadani\footnote{Vanderbilt University,
  \href{mailto:rahul.bhadani@vanderbilt.edu}{\nolinkurl{rahul.bhadani@vanderbilt.edu}},
  \href{mailto:rahulbhadani@email.arizona.edu}{\nolinkurl{rahulbhadani@email.arizona.edu}}}\\
  Jonathan Sprinkle\footnote{Vanderbilt University,
  \href{mailto:jonathan.sprinkle@vanderbilt.edu}{\nolinkurl{jonathan.sprinkle@vanderbilt.edu}}  }}
\date{}

\begin{document}
\maketitle
\begin{abstract}
This paper demonstrates the integration model-based design approaches
for vehicle control, with validation in a freely available open-source
simulator. 
Continued interest in autonomous vehicles and their deployment is driven
by the potential benefits of their use. However, it can be challenging
to transition new theoretical approaches into unknown simulation environments. 
Thus, it is critical for experts from other fields, whose insights may be 
necessary to continue to advance autonomy, to be able to create
control applications with the potential to transition to practice.
In this article, we will explain how to use the CAT
Vehicle simulator and ROS packages to create and test vehicle controllers. The
methodology of developing the control system in this article takes the
approach of model-based design using Simulink, and the ROS Toolbox, followed
by code generation to create a standalone C++ ROS node. Such ROS nodes
can be integrated through roslaunch in the CAT Vehicle ROS package.
\end{abstract}

\hypertarget{methods}{%
\section{Introduction}\label{methods}}

Recent advances in computing, control, and sensor technology have
brought autonomous systems -- especially autonomous ground vehicles (AV)
into the limelight of not only academic research but also media
\cite{zhu2020modelling}. The goal of autonomous vehicle control and
related research is to improve passenger comfort and safety
\cite{du2018velocity,mohajer2020enhancing} and reduce road
accidents\cite{michalowska2017autonomous}. At the same time, some
other researchers have been investigating the use of autonomous vehicle
control in reducing traffic congestion
\cite{delle2019feedback,stern2018dissipation,bhadani2019real,bhadani2018dissipation}
and fuel consumption reduction
\cite{lichtle2021fuel,qin2018stability}. Such objectives and
research endeavors encompass a mix of simulation study as well as
experimental research using physical platforms. While there are several
simulation software and packages have been developed to prototype
autonomous vehicle control -- both general-purpose simulators such as
AirSim \cite{shah2018airsim}, CARLA \cite{dosovitskiy2017carla};
and application-specific simulators such as CAT Vehicle
\cite{bhadani2018cat}, and \cite{wu2017flow}, not all simulators
are created equally. Some provide the ability to prototype a wide
variety of use cases but at the same arduous and difficult to get
familiar with, while others are limited in use cases but easier to
understand.

In this article, we present ways in which a previously proposed
autonomous vehicle simulator CAT Vehicle, written as a ROS package
\cite{quigley2009ros} can be used to prototype longitudinal vehicle
control. The CAT Vehicle simulator is a multi-vehicle simulator that
uses rigid body dynamics from the Gazebo physics engine
\cite{koenig2004design}. The methodology presented in this article
takes the approach of model-based design using Mathworks' Simulink.
Simulink provides ROS Toolbox that can be used to prototype ROS
components along with custom control law. Further, Simulink allows
code-generation of C++ standalone ROS node provides open-source C++
code, capable of executing on any Ubuntu machine.

The rest of the article is divided into the following parts. Section
\ref{sec:install_ros} provides a step-by-step guide to installing the
CAT Vehicle ROS package. Section \ref{sec:API} provides a brief overview
of CAT Vehicle APIs available for controller prototype. Section
\ref{sec:example} provides an example of vehicle control. We end the
article with a conclusion.

\hypertarget{installing-cat-vehicle-package}{%
\section{Installing CAT Vehicle
Package}\label{installing-cat-vehicle-package}}

\label{sec:install_ros} The CAT Vehicle is a ROS-based simulator written
as a ROS package to facilitate the development of autonomous vehicle
control applications. The simulator utilizes Gazebo 3D world and ROS
tools for deploying and testing a control application in a 3D
environment with realistic vehicle dynamics. In this section, we provide
a hands-on on how to install the CAT Vehicle ROS package that will help
you in getting started with writing an autonomous control application.
The examples presented in this article use ROS Noetic on Ubuntu 20.04.

\hypertarget{installing-ros-noetic}{%
\subsection{Installing ROS Noetic}\label{installing-ros-noetic}}

Open the terminal, and execute the following commands

\begin{Shaded}
\begin{Highlighting}[]
\FunctionTok{sudo}\NormalTok{ sh }\AttributeTok{{-}c} \StringTok{\textquotesingle{}echo "deb http://packages.ros.org/ros/ubuntu $(lsb\_release {-}sc) main" \textbackslash{} }
\StringTok{\textgreater{} /etc/apt/sources.list.d/ros{-}latest.list\textquotesingle{}}

\FunctionTok{sudo}\NormalTok{ apt install curl }\CommentTok{\# if you haven\textquotesingle{}t already installed curl}
\ExtensionTok{curl} \AttributeTok{{-}s}\NormalTok{ https://raw.githubusercontent.com/ros/rosdistro/master/ros.asc }\KeywordTok{|} \ExtensionTok{\textbackslash{} }
\FunctionTok{sudo}\NormalTok{ apt{-}key add }\AttributeTok{{-}}

\FunctionTok{sudo}\NormalTok{ apt{-}get update}

\FunctionTok{sudo}\NormalTok{ apt install ros{-}noetic{-}desktop{-}full}

\BuiltInTok{echo} \StringTok{"source /opt/ros/noetic/setup.bash"} \OperatorTok{\textgreater{}\textgreater{}}\NormalTok{ \textasciitilde{}/.bashrc}

\FunctionTok{sudo}\NormalTok{ apt install python3{-}rosdep python3{-}rosinstall }\DataTypeTok{\textbackslash{} }
\ExtensionTok{python3{-}rosinstall{-}generator}\NormalTok{ python3{-}wstool build{-}essential python3{-}rosdep}
\end{Highlighting}
\end{Shaded}

Once successfully executed above commands, close the terminal, reopen
it, and execute the following command

\begin{Shaded}
\begin{Highlighting}[]
\FunctionTok{sudo}\NormalTok{ rosdep init}
\ExtensionTok{rosdep}\NormalTok{ update}
\end{Highlighting}
\end{Shaded}

In addition, we require a few additional packages that can be installed
using the following command:

\begin{Shaded}
\begin{Highlighting}[]
\FunctionTok{sudo}\NormalTok{ apt{-}get install python{-}yaml}
\FunctionTok{sudo}\NormalTok{ apt{-}get install ros{-}noetic{-}controller{-}manager }\DataTypeTok{\textbackslash{} }
\ExtensionTok{ros{-}noetic{-}ros{-}control}\NormalTok{ ros{-}noetic{-}ros{-}controllers }\DataTypeTok{\textbackslash{} }
\ExtensionTok{ros{-}noetic{-}gazebo{-}ros{-}control}\NormalTok{ libpcap{-}dev ros{-}noetic{-}velodyne}
\end{Highlighting}
\end{Shaded}

\hypertarget{creating-catkin-workspace}{%
\subsection{Creating Catkin Workspace}\label{creating-catkin-workspace}}

The first step in using the CAT Vehicle ROS package is to create a
catkin workspace. Open a Terminal in your Ubuntu machine and type the
following:

\begin{Shaded}
\begin{Highlighting}[]
\BuiltInTok{cd}\NormalTok{ \textasciitilde{}}
\FunctionTok{mkdir} \AttributeTok{{-}p}\NormalTok{ catvehicle\_ws/src}
\BuiltInTok{cd}\NormalTok{ catvehicle\_ws/src}
\ExtensionTok{catkin\_init\_workspace}
\BuiltInTok{cd}\NormalTok{ ..}
\ExtensionTok{catkin\_make}
\end{Highlighting}
\end{Shaded}

Next, we will clone a few essential repositories that are dependencies
for the CAT Vehicle package

\begin{Shaded}
\begin{Highlighting}[]
\FunctionTok{git}\NormalTok{ clone https://github.com/jmscslgroup/catvehicle}
\FunctionTok{git}\NormalTok{ clone https://github.com/jmscslgroup/obstaclestopper}
\FunctionTok{git}\NormalTok{ clone https://github.com/jmscslgroup/control\_toolbox}
\FunctionTok{git}\NormalTok{ clone https://github.com/jmscslgroup/sicktoolbox}
\FunctionTok{git}\NormalTok{ clone https://github.com/jmscslgroup/sicktoolbox\_wrapper}
\FunctionTok{git}\NormalTok{ clone https://github.com/jmscslgroup/stepvel}
\FunctionTok{git}\NormalTok{ clone https://github.com/jmscslgroup/cmdvel2gazebo}
\BuiltInTok{cd}\NormalTok{ catvehicle}
\FunctionTok{git}\NormalTok{ checkout noetic\_gazebo{-}11}
\BuiltInTok{cd}\NormalTok{ \textasciitilde{}/catvehicle\_ws/}
\ExtensionTok{catkin\_make}
\end{Highlighting}
\end{Shaded}

\texttt{catkin\_make} compiles all packages and generates two folders in
\texttt{\textasciitilde{}/catvehicle\_ws} with the name \texttt{devel}
and \texttt{build}. They contain executables and other artifacts to run
the program written in ROS packages.

\hypertarget{sourcing-workspace-to-the-environment-path}{%
\subsection{Sourcing Workspace to the Environment
Path}\label{sourcing-workspace-to-the-environment-path}}

We also need to tell the terminal where to find the desired program we
want to run. For that, we need to ``source'' the \texttt{catvehicle\_ws}
catkin workspace. We do this by typing the following in the terminal:

\begin{Shaded}
\begin{Highlighting}[]
\BuiltInTok{echo} \StringTok{"source \textasciitilde{}/catvehicle\_ws/devel/setup.bash"} \OperatorTok{\textgreater{}\textgreater{}}\NormalTok{ \textasciitilde{}/.bashrc}
\BuiltInTok{source}\NormalTok{ \textasciitilde{}/.bashrc}
\end{Highlighting}
\end{Shaded}

Once done, close your terminal and reopen it. To test your installation,
type the following in one terminal

\begin{Shaded}
\begin{Highlighting}[]
\ExtensionTok{roslaunch}\NormalTok{ catvehicle catvehicle\_neighborhood.launch}
\end{Highlighting}
\end{Shaded}

and the following in the second terminal:

\begin{Shaded}
\begin{Highlighting}[]
\ExtensionTok{gzclient}
\end{Highlighting}
\end{Shaded}

\begin{figure}[htbp]
\centering
\includegraphics[width=0.4\textwidth]{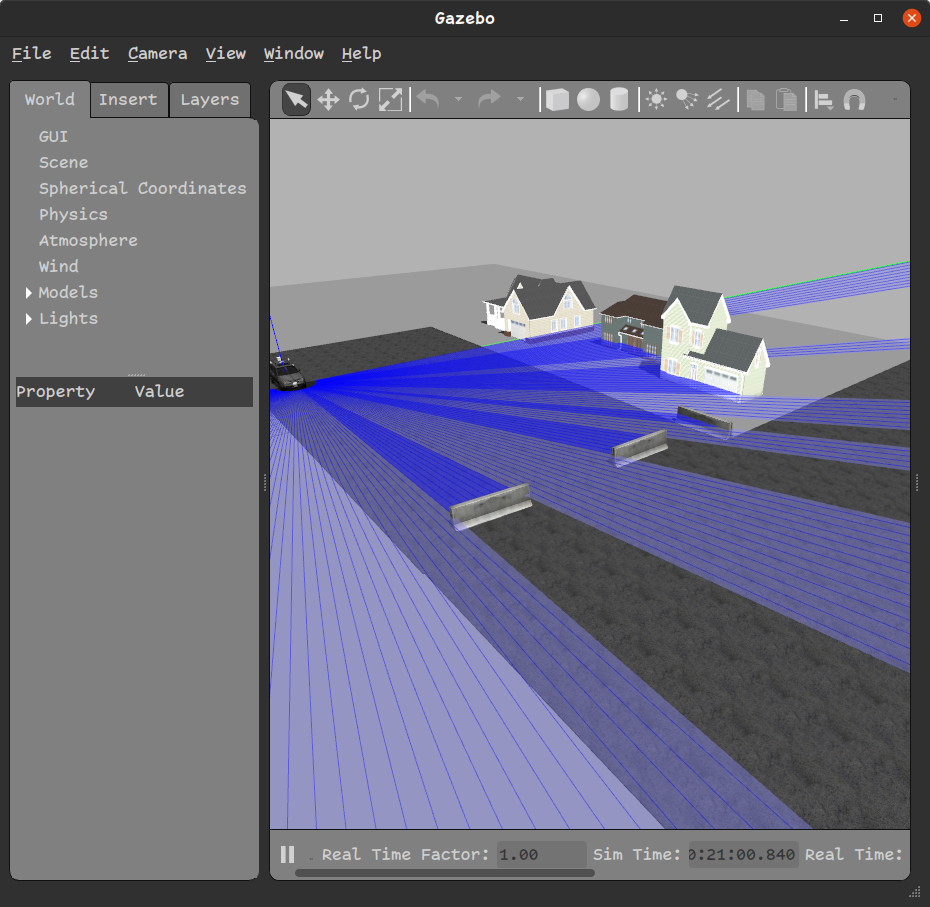}
\caption{A Gazebo window with an example simulated environment.}
\label{fig:Gazebo_001.png}
\end{figure}

\texttt{gzclient} should open a Gazebo window that should look like the
one shown in Figure \ref{fig:Gazebo_001.png}.

\hypertarget{cat-vehicle-apis-for-vehicle-control-applications}{%
\section{CAT Vehicle APIs for Vehicle Control
Applications}\label{cat-vehicle-apis-for-vehicle-control-applications}}

\label{sec:API} While this section is not a tutorial on how to use ROS,
it is necessary to understand a few basic things about ROS that can help
create some simple control applications. we explain the basics using
what is provided through the CAT Vehicle package.

\hypertarget{the-launch-file}{%
\subsection{The Launch File}\label{the-launch-file}}

ROS provides a methodology to execute a specialized program called ROS
nodes through launch files. ROS nodes do some meaningful tasks (such as
executing a control law) and publish messages on a named topic or
subscribe to some other messages through a named topic. At the same
time, some other ROS nodes can subscribe to messages being published
through topics. Topics are like slots and nodes put messages on those
slots -- some other node can read those slots to get messages.

Launch files have an extension \texttt{.launch} and they are generally
in the launch directory of a ROS package. In the case of the CAT Vehicle
package, consider the launch file \texttt{catvehicle\_empty.launch}. It
can be used to create a simulation by typing the following in a terminal

\begin{Shaded}
\begin{Highlighting}[]
\ExtensionTok{roslaunch}\NormalTok{ catvehicle catvehicle\_empty.launch}
\end{Highlighting}
\end{Shaded}

To see the visual, we type the following in another terminal:

\begin{Shaded}
\begin{Highlighting}[]
\ExtensionTok{gzclient}
\end{Highlighting}
\end{Shaded}

\begin{figure}[htbp]
\centering
\includegraphics[width=0.4\textwidth]{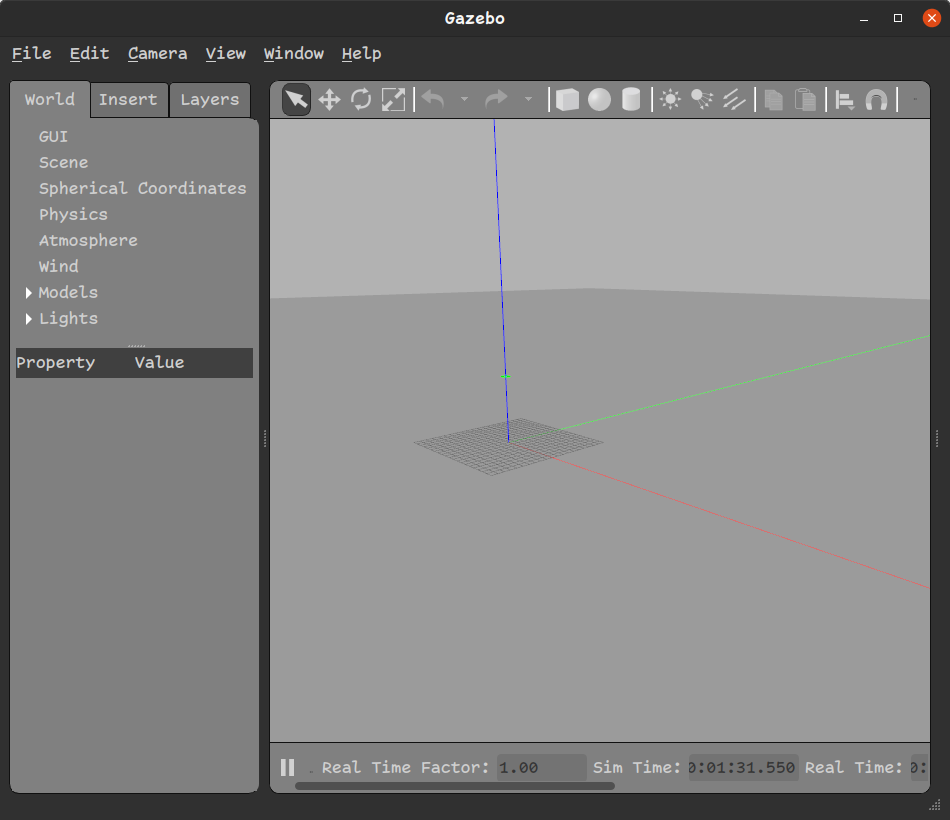}
\caption{An empty Gazebo window.}
\label{fig:Gazebo_002.png}
\end{figure}

The above command launches a window in the Gazebo program showing a
virtual world with a ground plane and the center coordinates as shown in
Figure \ref{fig:Gazebo_002.png}.

\hypertarget{spawning-a-vehicle}{%
\subsection{Spawning a Vehicle}\label{spawning-a-vehicle}}

Spawning a vehicle in the virtual world is done using the
\texttt{catvehicle\_spawn.launch} file in another terminal.

\begin{Shaded}
\begin{Highlighting}[]
\ExtensionTok{roslaunch}\NormalTok{ catvehicle catvehicle\_spawn.launch}
\end{Highlighting}
\end{Shaded}

By default, it creates a vehicle at the origin with the name
\texttt{catvehicle}. The launch file provides several command line
arguments that can be revealed by pressing the tab a couple of times
after typing \texttt{roslaunch\ catvehicle\ catvehicle\_spawn.launch} in
the terminal. We have the following command line arguments:

\begin{itemize}
\tightlist
\item
  camera\_left: to enable the left camera mounted on the car.
\item
  camera\_right: to enable the right camera mounted on the car.
\item
  laser\_sensor: to enable front 2-D Lidar sensor.
\item
  obstaclestopper: to enable a custom control node that prevents
  collision.
\item
  pitch: specify the pitch in radian.
\item
  robot: the name of the car. you must specify a unique name when
  spawning multiple cars in the simulation.
\item
  roll: specify the roll in radian.
\item
  triclops: enable front-mounted camera on the car.
\item
  updateRate: specify the update rate of speed of the car published.
\item
  velodyne\_points: enable 3D Velodyne Lidar sensor.
\item
  X: specify the X coordinate of the car.
\item
  Y: specify the Y coordinate of the car.
\item
  yaw: specify the yaw of the car in radian.
\item
  Z: specify the Z coordinate of the car.
\end{itemize}

With some of the most essential options, we can spawn a car with the
following command-line arguments:

\begin{Shaded}
\begin{Highlighting}[]
\ExtensionTok{roslaunch}\NormalTok{ catvehicle spawn.launch robot:=ego X:=0.0 laser\_sensor:=true}
\end{Highlighting}
\end{Shaded}

\begin{figure}[htbp]
\centering
\includegraphics[width=0.4\textwidth]{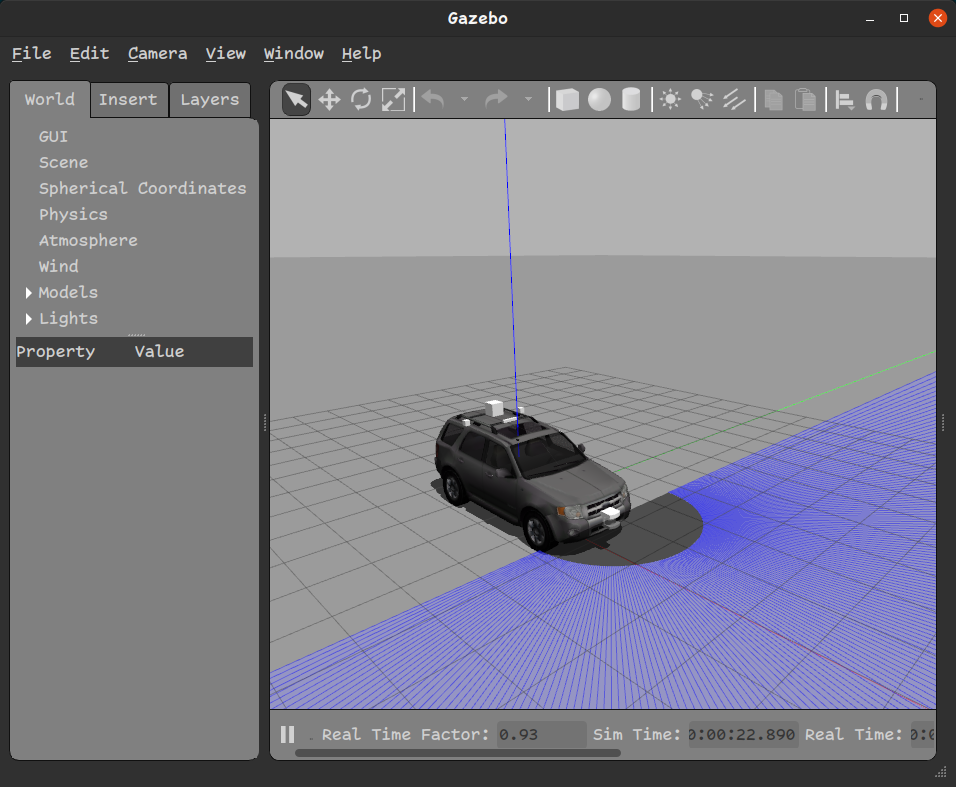}
\caption{A car spawned at the center with a 2-D laser sensor}
\label{fig:Gazebo_003.png}
\end{figure}

The above command spawns a car at the center with the name \texttt{ego}
and a front 2D Lidar sensor enabled. Figure \ref{fig:Gazebo_003.png}
displays the outcome.

\hypertarget{important-rostopics-in-the-cat-vehicle-package}{%
\subsection{Important ROStopics in the CAT Vehicle
package}\label{important-rostopics-in-the-cat-vehicle-package}}

To develop a control application, we will need to know about some
important ROS topics. A full list of topics can be obtained by typing
\texttt{rostopic\ list} and anything that starts with \texttt{/ego} are
topics associated with the above car we spawned. \texttt{/ego/vel} is
where we get the current driving speed of the car on its
\texttt{linear.x} component. Note that each topic has a message type
that is equivalent to a C++ structure. You can see the message types of
each topic in the output of the \texttt{rostopic\ list}. Interested
readers can learn more about ROS messages at
\url{http://wiki.ros.org/msg}. The relative speed of any car being
followed by a car directly in its front can be found on the
\texttt{linear.z} component of \texttt{/ego/rel\_vel}. It will be zero
if there is no car in the front. Headway distance of the leader car in
front of the ego car is obtained on the topic \texttt{/ego/lead\_dist}
on the \texttt{data} component. A control command to the car can be sent
on the topic \texttt{/ego/cmd\_vel} where you can specify speed on the
\texttt{linear.x} component and steering angle on the \texttt{angular.z}
component.

\hypertarget{controller-modeling-example}{%
\section{Controller Modeling
Example}\label{controller-modeling-example}}

\label{sec:example}

For controller modeling, we take the approach of model-based design
using Simulink software which is a part of Mathworks' MATLAB. Simulink
provides a library of blocks for specific purposes. One such blocks are
ROS toolbox that can be used for creating controller models.

\hypertarget{modeling-in-simulink}{%
\subsection{Modeling in Simulink}\label{modeling-in-simulink}}

Open MATLAB, in the MATLAB command prompt, type \texttt{simulink}.
Select ``Create Model'' in the \textbf{Blank Model} option. In the empty
model workspace, you can see \textbf{Library Browser} where you can
choose, drag-and-drop blocks to perform certain tasks. We are interested
in blocks from \textbf{ROS Toolbox}. Note that my example is built in
MATLAB 2022b. We are interested in a naive velocity control shown in
Equation 1 which we arbitrarily came up with. This control law is merely
for following a vehicle in its front if there is one. \begin{equation}
\label{eq:vel_control}
v_{\textrm{cmd}} = \begin{cases}r + 0.5 v_{\textrm{lead}}\quad ~\textrm{if}~ h > 30 \\
r \quad ~\textrm{if}~ h = 30\\
r - 0.5 v_{\textrm{lead}}\quad ~\textrm{if}~ h < 30
\end{cases}
\end{equation}

In Equation \eqref{eq:vel_control}, \(r\) is the desired velocity for
the ego vehicle (The ego vehicle is the one we are interested in
controlling). \(v_{\textrm{lead}}\) is the speed of the vehicle or an
object directly in the range of the ego vehicle's front LiDAR sensor.
\(v_{\textrm{lead}}\) is reconstructed from LiDAR data by
differentiating headway \(h\) (that is available on the
\texttt{/ego/lead\_dist} topic) and adding to the ego's current velocity
\(v\) (obtained from the topic \texttt{/ego/vel}). Differentiated
relative velocity is published on \texttt{/ego/rel\_veltopic}.
\(v_{\textrm{cmd}}\) is published on the topic \texttt{/ego/cmd\_vel}.
Note that \texttt{/ego/cmd\_vel} and \texttt{/ego/vel} are different
because a vehicle has dynamics so it won't exactly be driving with what
it is commanded to do so. We have a hidden transfer function to
represent the vehicle dynamics but we don't model it separately. It is
done by rigid body dynamics implemented in the CAT Vehicle package.

\begin{figure}[htbp]
\centering
\includegraphics[width=1.0\textwidth]{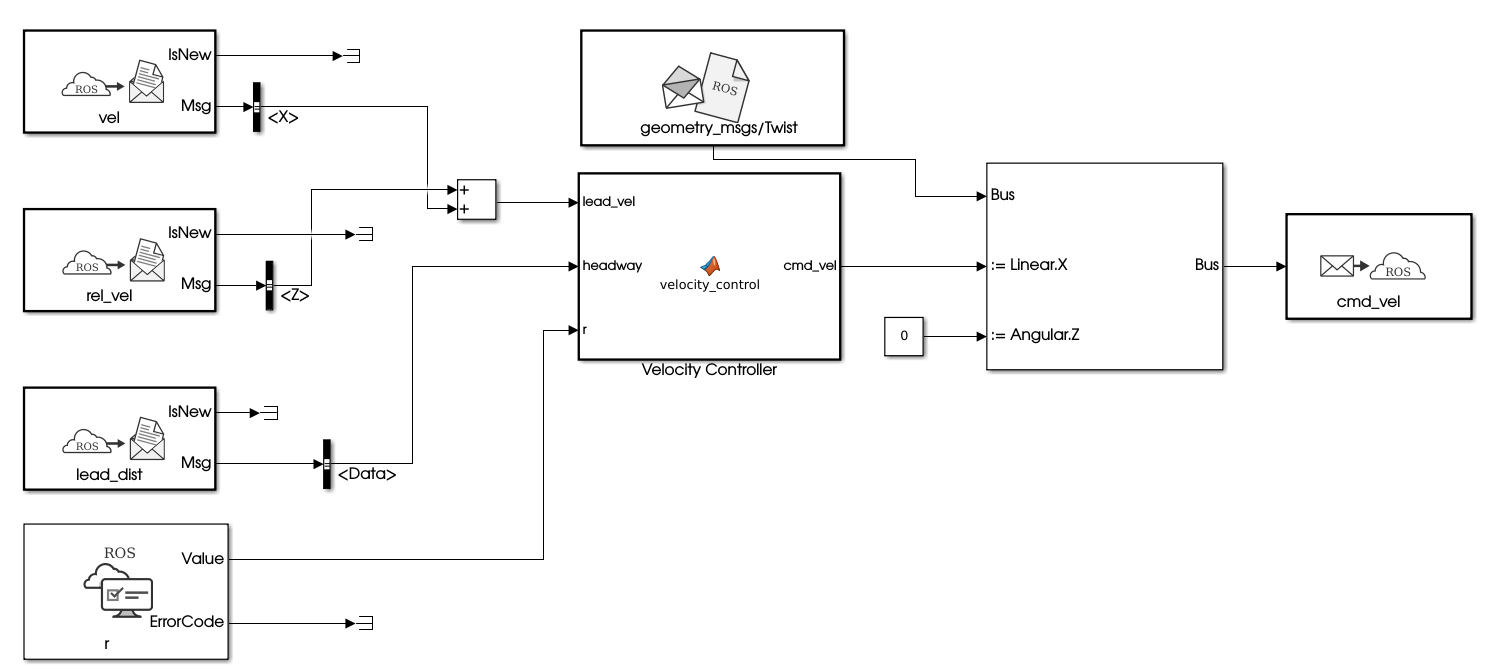}
\caption{The Simulink model of the velocity controller}
\label{fig:simulink_model.png}
\end{figure}

A full model of Equation \eqref{eq:vel_control} in Simulink is shown in
Figure \ref{fig:simulink_model.png} where the Equation is contained in
MATLAB function block.

\hypertarget{settings-for-the-model}{%
\subsection{Settings for the model}\label{settings-for-the-model}}

In the Simulink \textbf{Simulation} tab, we set the stop time of the
simulation to be \texttt{inf}. Now, we specify ROS-related parameters in
the \textbf{Modeling} tab -\textgreater{} \textbf{Model Settings} to
generate a ROS node. In \textbf{Model Settings}, we use the following
settings:

\begin{enumerate}
\def\labelenumi{\arabic{enumi}.}
\tightlist
\item
  Solver -\textgreater{} Type: Fixed Step, Fixed-Step Size: 0.05 (which
  is in seconds)
\item
  Hardware implementation-\textgreater{} Hardware Board: Robot Operating
  System, Target Hardware Resources-\textgreater{} Build Options: Build
  and Load, Catkin Workspace: \texttt{\textasciitilde{}/catvehicle\_ws/}
  (or \texttt{/home/\textless{}username\textgreater{}/catvehicle\_ws/})
\end{enumerate}

Then we press \textbf{OK}. We save the model as
\texttt{velocity\_control.slx}. The Simulink file used in this example
can be downloaded from
\url{https://github.com/rahulbhadani/medium.com/blob/master/10-30-2022/velocity_control.slx}.

\hypertarget{generating-ros-node-and-corresponding-launchfile-from-the-simulink-model}{%
\subsection{Generating ROS node and corresponding launchfile from the
Simulink
model}\label{generating-ros-node-and-corresponding-launchfile-from-the-simulink-model}}

To generate the ROS node, we type \texttt{roscore} in a terminal window,
and then in the Simulink ROS tab, press \textbf{Build \& Load}. It
compiles the model and generates a C++ standalone ROS node in
\texttt{\textasciitilde{}/catvehicle\_ws/src}. The first step in running
the simulation is to create a launch file. We first create a new text
file in an editor and copy the following code:

\begin{Shaded}
\begin{Highlighting}[]
\OperatorTok{\textless{}}\ExtensionTok{?xml}\NormalTok{ version=}\StringTok{"1.0"}\NormalTok{ encoding=}\StringTok{"UTF{-}8"}\PreprocessorTok{?}\OperatorTok{\textgreater{}}
\OperatorTok{\textless{}}\NormalTok{launch}\OperatorTok{\textgreater{}}
\OperatorTok{\textless{}}\NormalTok{arg }\VariableTok{name}\OperatorTok{=}\StringTok{"robot"} \VariableTok{default}\OperatorTok{=}\StringTok{"ego"}\NormalTok{/}\OperatorTok{\textgreater{}}
    \OperatorTok{\textless{}}\NormalTok{arg }\VariableTok{name}\OperatorTok{=}\StringTok{"r"} \VariableTok{default}\OperatorTok{=}\StringTok{"20.0"}\NormalTok{/}\OperatorTok{\textgreater{}}
    \OperatorTok{\textless{}}\NormalTok{param }\VariableTok{name}\OperatorTok{=}\StringTok{"/}\VariableTok{$(}\ExtensionTok{arg}\NormalTok{ robot}\VariableTok{)}\StringTok{/r"} \VariableTok{type}\OperatorTok{=}\StringTok{"double"} \VariableTok{value}\OperatorTok{=}\StringTok{"}\VariableTok{$(}\ExtensionTok{arg}\NormalTok{ r}\VariableTok{)}\StringTok{"}\NormalTok{/}\OperatorTok{\textgreater{}}
    \OperatorTok{\textless{}}\NormalTok{group }\VariableTok{ns}\OperatorTok{=}\StringTok{"ego"}\OperatorTok{\textgreater{}}
        \OperatorTok{\textless{}}\NormalTok{node }\VariableTok{pkg}\OperatorTok{=}\StringTok{"velocity\_control"} \VariableTok{type}\OperatorTok{=}\StringTok{"velocity\_control"} 
                \VariableTok{name}\OperatorTok{=}\StringTok{"velocity\_control\_node"} \VariableTok{output}\OperatorTok{=}\StringTok{"screen"}\NormalTok{/}\OperatorTok{\textgreater{}}
     \OperatorTok{\textless{}}\NormalTok{/group}\OperatorTok{\textgreater{}}
     
\OperatorTok{\textless{}}\NormalTok{/launch}\OperatorTok{\textgreater{}}
\end{Highlighting}
\end{Shaded}

We save the text file as \texttt{velocity\_control.launch} in the launch
folder of the \texttt{catvehicle} package (which may be in the
\texttt{\textasciitilde{}/catvehicle\_ws/src/catvehicle/launch}
directory).

\hypertarget{simulation-setup}{%
\subsection{Simulation Setup}\label{simulation-setup}}

We consider a two-vehicle simulation where the first vehicle or the
leader vehicle drives with an open loop trajectory specified from a data
file. The data file can be downloaded from
\url{https://github.com/rahulbhadani/medium.com/releases/download/data/test_data.csv}
The leader vehicle control is executed using
\texttt{velinjector.launch}. For the purpose of this tutorial, we save
data in the home directory. A whole setup is illustrated in Figure 4.

\begin{figure}[htbp]
\centering
\includegraphics[width=0.5\textwidth]{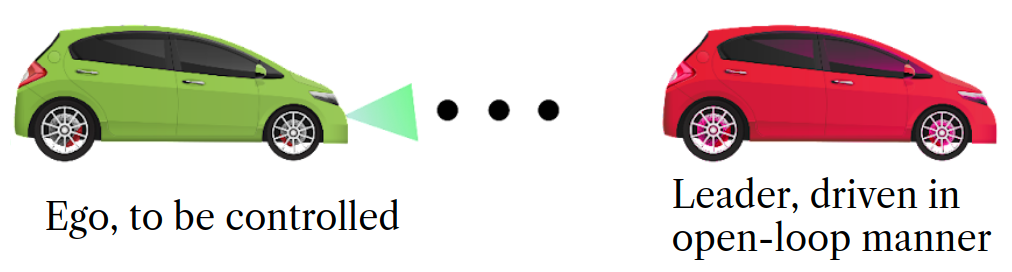}
\caption{Two-car simulation setup for the velocity control example}
\label{fig:ego_leader.png}
\end{figure}

\hypertarget{running-the-simulation}{%
\subsection{Running the Simulation}\label{running-the-simulation}}

To run the simulation with our velocity controller developed in the
Simulink, we need to execute several \texttt{roslaunch} files in
different terminal windows. To make things easier, we can use the bash
script below which executes all \texttt{roslaunch} one by one.

\begin{Shaded}
\begin{Highlighting}[]
\CommentTok{\#!/bin/bash}
\ExtensionTok{gnome{-}terminal} \AttributeTok{{-}{-}}\NormalTok{ roslaunch catvehicle catvehicle\_empty.launch}
\FunctionTok{sleep}\NormalTok{ 5}
\ExtensionTok{gnome{-}terminal} \AttributeTok{{-}{-}}\NormalTok{ roslaunch catvehicle catvehicle\_spawn.launch robot:=leader X:=30.0}
\FunctionTok{sleep}\NormalTok{ 5}
\ExtensionTok{gnome{-}terminal} \AttributeTok{{-}{-}}\NormalTok{ gzclient}
\FunctionTok{sleep}\NormalTok{ 5}
\ExtensionTok{gnome{-}terminal} \AttributeTok{{-}{-}}\NormalTok{ roslaunch catvehicle spawn.launch robot:=ego X:=0.0 laser\_sensor:=true }
\FunctionTok{sleep}\NormalTok{ 5}
\VariableTok{velinjectfile}\OperatorTok{=}\StringTok{"roslaunch catvehicle velinjector.launch}
\StringTok{csvfile:=/home/ubuntu/test\_data.csv input\_type:=CSV }
\StringTok{time\_col:=Time vel\_col:=speed robot:=leader str\_angle:=0.0"}
\ExtensionTok{gnome{-}terminal} \AttributeTok{{-}{-}} \VariableTok{$velinjectfile}
\FunctionTok{sleep}\NormalTok{ 5}
\ExtensionTok{gnome{-}terminal} \AttributeTok{{-}{-}}\NormalTok{ roslaunch catvehicle velocity\_control.launch robot:=ego r:=2.5}
\FunctionTok{sleep}\NormalTok{ 5}
\ExtensionTok{gnome{-}terminal} \AttributeTok{{-}{-}}\NormalTok{ rosparam set /execute true}
\end{Highlighting}
\end{Shaded}

We save the above bash script as \texttt{run\_controller.sh} and execute
the following to run the simulation

\begin{Shaded}
\begin{Highlighting}[]
\FunctionTok{chmod}\NormalTok{ +x run\_controller.sh}
\ExtensionTok{./run\_controller.sh}
\end{Highlighting}
\end{Shaded}

The above command opens a series of terminal windows and executes all
commands one by one.

To log the data in a \texttt{.bag} format, type
\texttt{rosbag\ record\ -a}. The \texttt{.bag} file can be analyzed
using the \texttt{bagpy} python package. How to use the \texttt{bagpy}
package can be found at \url{https://jmscslgroup.github.io/bagpy}. To
terminate the simulation, we press \texttt{Ctrl-C} in every terminal
window that was opened through the bash script. To stop the rosbag
recording, we also need to press \texttt{Ctrl-C}.

\hypertarget{conclusion-and-discussion}{%
\section{Conclusion and Discussion}\label{conclusion-and-discussion}}

In this article, we have discussed how to use the CAT Vehicle ROS
package and Simulink's model-based design approach to prototype a
vehicle control law and test it in the Simulation. The example presented
in this article uses local data corresponding to the ego vehicle,
however, a more complex control law that uses non-local data is also
possible. Further other sensor information such as front and side-camera
can be used for improving the decision-making ability of the velocity
control. However, based on the current implementation of the simulator
in the CAT Vehicle package, only a velocity control command is possible.
If acceleration-based control law needs to be prototyped, one needs to
take an indirect approach of adding an integrator block in Simulink to
integrate the commanded acceleration to produce a velocity command.

\bibliographystyle{plain}
 \bibliography{IEEEabrv,biblio}
 
\end{document}